\documentclass{Interspeech}

\interspeechcameraready

\usepackage{svg}
\usepackage{multirow}
\usepackage{graphicx}
\usepackage{arydshln,subcaption}
\usepackage{hyperref}

\title{Jointly Improving Dialect Identification and ASR in Indian Languages using Multimodal Feature Fusion}

\author{Saurabh}{Kumar}
\author{Amartyaveer}{}
\author{Prasanta Kumar}{Ghosh}

\affiliation[nocounter]{Department of Electrical Engineering}{Indian Institute of Science, Bangalore}{India}

\email{saurabhk0317@gmail.com, amartyaveer72@gmail.com, prasantag@gmail.com}

\usepackage{comment}

\begin{document}

\maketitle

\begin{abstract}
Automatic Speech Recognition (ASR) and Dialect Identification (DID) are crucial for Indian languages, many of which are low-resource and exhibit significant dialectal differences. Existing methods often optimize ASR or DID individually, resulting in performance trade-offs. In this work, we propose a multimodal framework that jointly improves ASR and DID. Our method employs a Bottleneck Encoder to extract dialectal features from Conformer-based speech representations and a RoBERTa encoder to process ASR-generated CTC embeddings. A gating mechanism merges these features, followed by an attention encoder to refine the representations. The learned embeddings are concatenated with Conformer outputs to enhance ASR features. Evaluated on eight Indian languages with thirty-three dialects, our method achieves an average DID accuracy of $81.63\%$ and average CER and WER of $4.65\%$ and $17.73\%$, respectively. These results highlight the effectiveness of our method for joint ASR-DID modeling.
\end{abstract}

\section{Introduction}


Automatic Speech Recognition (ASR) and Dialect Identification (DID) are crucial for advancing speech technology, particularly in linguistically diverse regions like India. ASR systems rely on robust acoustic and linguistic models to transcribe speech accurately, while DID facilitates the adaptation of ASR models to different dialectal variations. Accurately identifying dialects enhances ASR performance by mitigating errors stemming from pronunciation, vocabulary, and grammatical variations \cite{Udupa_ASRU2023}. Despite their interdependence, ASR and DID have traditionally been studied as separate tasks, leading to suboptimal performance in dialect-sensitive ASR applications.

DID is inherently more challenging than Language Identification (LID). While LID involves distinguishing between different languages, DID requires differentiating between dialects of the same language, which often share similar phonetic, lexical, and syntactic characteristics \cite{singh2021spoken,montavon2009deep,punjabi2021joint,lyu2022ant}. These similarities make DID a significantly more difficult classification task than LID. Many prior works have explored joint ASR and LID using multi-task learning to enhance system performance \cite{punjabi2021joint,zhang22da_interspeech,William_LID_ICASSP2023,Zhou_ICASSP2022}, but fewer studies have focused on joint ASR and DID \cite{Imaizumi_Japanese_DID2022,lonergan2024low,ICASSP25_DID}. Research on joint ASR-DID in Indian languages remains limited \cite{ICASSP25_DID}, though recent studies have introduced dialectal ASR systems for these languages \cite{Bhardwaj_Punjabi_ASR_2021,Priya_MLASR2022,Arunkumar2022DuDeDM,Amitoj_ASR_Survey_2020,diwan21_interspeech,Udupa_ASRU2023,ICASSP25_DID}.

While joint ASR-DID methods \cite{Imaizumi_Japanese_DID2022,lonergan2024low,ICASSP25_DID} significantly outperform traditional DID approaches relying solely on audio or text-based features \cite{luo23c_interspeech,sullivan23_interspeech,kakouros23_interspeech,shaik_24_icassp,radhakrishnan23_interspeech,mishra-mujadia-2019-arabic,Althobaiti_text_DID_2020,goswami2020unsupervised}, ASR systems in such setups often perform suboptimally. In many cases, improvements in DID come at the cost of degraded ASR performance. For example, \cite{Imaizumi_Japanese_DID2022} reported that a multi-dialect Japanese ASR system improved DID accuracy by treating it as an auxiliary task. However, the best ASR and DID performances were achieved in different setups, with the best DID model degrading ASR performance, particularly in cases of incorrect DID predictions. Similarly, \cite{lonergan2024low} proposed an ASR-based DID system for Irish dialects using an intermediate CTC loss, which improved DID performance but caused slight degradation in ASR performance. This trade-off highlights the necessity for a joint modeling approach that effectively balances both tasks.

Recently, \cite{ICASSP25_DID} proposed an ASR-based DID approach using multimodal features from a dialect-aware ASR model jointly trained in a multi-task setup, achieving state-of-the-art DID accuracies in eight Indian languages. However, no improvement in ASR performance was reported, likely due to inherent limitations such as the lack of gradient propagation from the DID block to the ASR block and the prepending of text with dialect tokens, which may lead the ASR model to learn false context when predicted dialect IDs are incorrect.

Motivated by these challenges, we propose a novel framework for joint ASR-DID in a multi-task setup using multimodal feature fusion~\footnote{Code and models publicly available at: \url{https://github.com/labspire/respin_did_interspeech25.git}}. Our approach employs a Bottleneck Encoder to capture dialectal variations in Conformer-based speech representations and a RoBERTa encoder to extract dialectal cues from CTC embeddings derived from Conformer output. These complementary representations are fused using a gating mechanism, followed by an attention encoder to refine dialectal representations. To further enhance ASR performance, the learned attention embeddings are concatenated with the Conformer encoder output, enabling the extraction of richer ASR features through additional attention layers. Unlike previous methods, our proposed ASR-DID framework does not prepend training texts with dialect IDs, resulting in significant ASR performance improvements for utterances with incorrect dialect predictions.

We evaluate our proposed method on eight Indian languages, comprising $33$ dialects from various regions of India. Experimental results demonstrate significant improvements in both ASR and DID compared to existing state-of-the-art methods. The proposed framework effectively mitigates the trade-off between ASR and DID by leveraging multimodal attention fusion to achieve robust joint modeling. This work contributes to the broader goal of developing more accurate and adaptable speech technologies for Indian languages, particularly in linguistically diverse settings.

\section{Proposed Method}
\label{sec:proposed}

\begin{figure}[!t]
    \centering
    \begin{subfigure}{0.45\columnwidth}
        \centering
        \includegraphics[height=5cm]{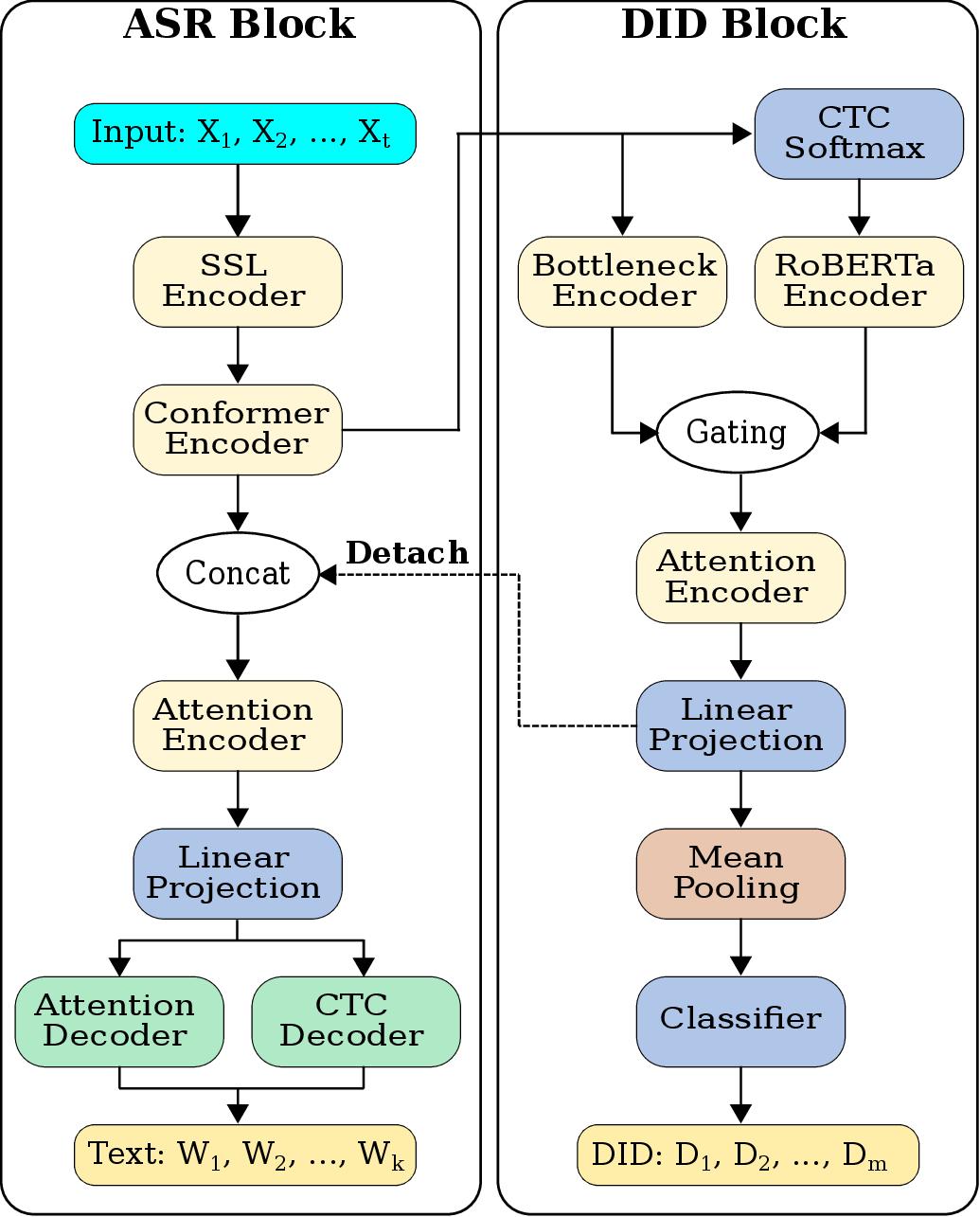} 
        \caption{}
        \label{fig:proposed_model}
    \end{subfigure}
    \hspace{1mm} 
    \begin{subfigure}{0.45\columnwidth}
        \centering
        \includegraphics[height=5cm]{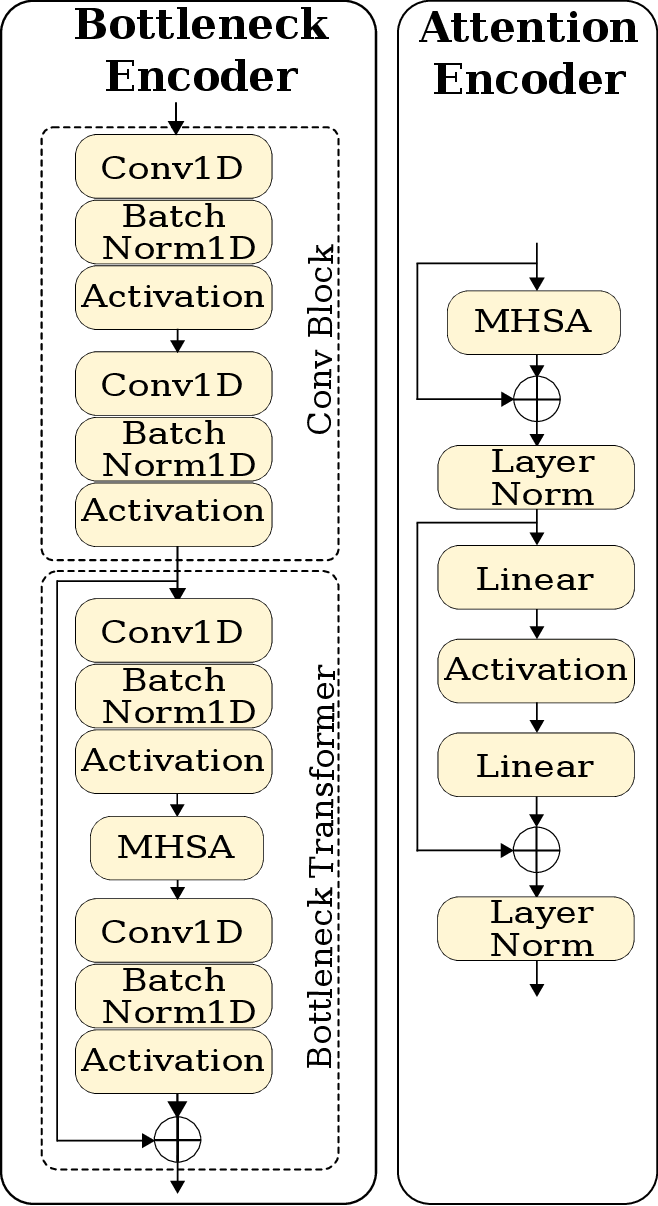} 
        \caption{}
        \label{fig:bneck_enc}
    \end{subfigure}\vspace{-3mm}
    \caption{Illustration of the (a) proposed architecture and (b) bottleneck encoder (left) and attention encoder (right).}\vspace{-5mm}
    \label{fig:combined_architecture}
\end{figure}

Our proposed method employs multimodal feature fusion to jointly enhance Automatic Speech Recognition (ASR) and Dialect Identification (DID) for Indian languages. As shown in Figure~\ref{fig:proposed_model}, the architecture consists of two primary components: the ASR Block and the DID Block, integrating speech and text-based dialectal cues for effective representation learning.

\subsection{ASR Block}
The ASR Block processes input speech using a self-supervised learning (SSL) encoder to extract phonetic and prosodic features, which are then refined by a Conformer Encoder. This module captures local speech patterns via convolutional layers and long-range dependencies through self-attention.

The Conformer output is directed to the DID Block, where multimodal representations are obtained by fusing features from a Bottleneck Encoder and a RoBERTa Encoder. The RoBERTa Encoder processes CTC embeddings, derived by projecting the Conformer output through a linear layer to the vocabulary size, followed by softmax (depicted by \textit{CTC Softmax} block in Figure~\ref{fig:proposed_model}). A gating mechanism combines these representations, which are further refined by an Attention Encoder and linearly projected to extract frame-level dialect embeddings.

To integrate multimodal features, dialect embeddings from the DID Block are concatenated with the Conformer output in the ASR Block. To prevent gradient interference, these embeddings are detached from the computational graph before concatenation, ensuring gradients flow only within the ASR Block.

The final combined representation is processed by an Attention Encoder and a Linear Projection before being fed into two parallel decoders: (1) an \textit{Attention Decoder} for context-aware token predictions and (2) a \textit{CTC Decoder} for non-autoregressive transcription. Both decoders contribute to generating the final ASR output.

\subsection{DID Block}
The DID Block is designed to classify dialects by integrating both speech and textual features. It takes the Conformer Encoder output as input and processes it through multiple specialized components.

\subsubsection{Bottleneck Encoder}
The Bottleneck Encoder extracts and refines speech-based dialectal features. It consists of a convolutional block for feature extraction, followed by a bottleneck transformer~\cite{Srinivas_bneck_2021} for global representation learning. As illustrated in Figure~\ref{fig:bneck_enc}(left), this architecture is inspired by~\cite{Alhakeem_bneck_did2024}, which employs $2D$ convolutional layers on both sides of multi-head self-attention (MHSA) with adaptive average pooling. However, to better preserve temporal information, we replace $2D$ convolutions with $1D$ convolutions and remove the average pooling layer. This refinement preserves critical temporal dynamics in ASR-encoded speech representations for dialect classification.

\subsubsection{RoBERTa Encoder}
Inspired by~\cite{Liu2019RoBERTaAR}, this module extracts contextual dialectal features from CTC embeddings. These embeddings are first linearly projected into a lower-dimensional space before being processed by RoBERTa, complementing speech-based features from the Bottleneck Encoder.

\subsubsection{Gating Mechanism}
To dynamically balance the contributions of speech and text-based features, a gating mechanism is employed. Given two feature representations, an adaptive weight function computes a set of learnable weights using a linear transformation followed by a sigmoid activation:

\begin{equation}
\mathbf{G} = \sigma(\mathbf{W}_\text{g} [\mathbf{H}_{\text{bottleneck}}, \mathbf{H}_{\text{rob}}] + \mathbf{b}_\text{g})
\end{equation}
where $\sigma$ represents the sigmoid activation, and $\mathbf{W}_\text{g}$ and $\mathbf{b}_\text{g}$ are learnable parameters. The concatenated feature representations are denoted as $[\mathbf{H}_{\text{bottleneck}}, \mathbf{H}_{\text{rob}}]$. The final fused representation is computed as:
\begin{equation}
\mathbf{H}_{\text{fused}} = \mathbf{G} \odot \mathbf{H}_{\text{rob}} + (1 - \mathbf{G}) \odot \mathbf{H}_{\text{bottleneck}}
\end{equation}
where $\odot$ denotes element-wise multiplication.

\subsubsection{Attention Encoder}
The fused representation is further refined through an Attention Encoder, which enhances contextual dependencies via multi-head self-attention (MHSA), followed by Layer Normalization. This module includes up-projection and down-projection layers with activation functions to refine the feature representations while employing residual connections for stable learning. The Attention Encoder shares its architecture with the one used in the ASR Block, as shown in Figure~\ref{fig:bneck_enc}(right). The final output is linearly projected to obtain frame-level dialect embeddings.

Mean pooling is applied to obtain a fixed-length representation, which is passed through a classifier with softmax activation for dialect prediction. By integrating speech and text-based features via adaptive gating and attention mechanisms, the DID Block effectively captures linguistic and acoustic nuances for robust dialect identification.

\subsection{Objective Function}
To jointly optimize ASR and DID tasks, we employ a weighted loss function. The total loss, $\mathcal{L}$, is a combination of the CTC loss ($\mathcal{L}_{\text{CTC}}$), attention loss ($\mathcal{L}_{\text{ATT}}$), and cross-entropy loss ($\mathcal{L}_{\text{CE}}$), formulated as:
\begin{equation}
\mathcal{L} = \lambda_{\text{CTC}} \mathcal{L}_{\text{CTC}} + (1 - \lambda_{\text{CTC}}) \mathcal{L}_{\text{ATT}} + \gamma_{\text{CE}} \mathcal{L}_{\text{CE}},
\end{equation}
where $\lambda_{\text{CTC}}$ manages the balance between CTC and attention-based ASR losses, and $\gamma_{\text{CE}}$ adjusts the influence of the cross-entropy loss for DID.

Our framework integrates ASR and DID with multimodal speech-text representations. Gating balances features, and detachment prevents gradient interference, ensuring stable optimization. This enhances transcription and dialect classification, making it ideal for India's diverse languages.


\section{Experiments}
\label{sec:experiments}
\subsection{Datasets}
We use a subset of the RESPIN dataset covering eight Indian languages: Bhojpuri (bh), Bengali (bn), Chhattisgarhi (ch), Kannada (kn), Magahi (mg), Maithili (mt), Marathi (mr), and Telugu (te). The train-test splits follow \cite{ICASSP25_DID}\footnote{The RESPIN dataset is publicly available at: \url{https://spiredatasets.ee.iisc.ac.in/respincorpus}}, with approximately $140-175$ hours of training data, $2$ hours of development data, and $6-8$ hours of test data per language, all in a read-speech setting.

\subsection{Experimental Setup}
Our experiments, conducted in ESPnet\footnote{ESPnet: \url{https://github.com/espnet/espnet.git}}, compare the proposed approach with state-of-the-art ASR-DID frameworks. The hybrid CTC/Attention ASR model comprises an $8$-block Conformer encoder ($256$-dim output, $4$ heads, $1024$-dim feedforward) and a $6$-block Transformer decoder ($4$ heads, $2048$-dim feedforward), both with a $0.1$ dropout. A CTC weight of $0.3$, cross-entropy scaling of $5$, and label smoothing of $0.1$ are applied. Adam optimization is used with a $0.002$ learning rate, $1\times10^{-6}$ weight decay, and a $1.5\times10^4$-step warmup. The batch size dynamically adjusts to $6M$ batch bins, with early stopping patience of $5$. The top $5$ models are selected based on validation accuracy. The frontend employs a pre-trained IndicWav2Vec model~\cite{Javed_indicw2v2_2021}\footnote{IndicWav2Vec: \url{https://github.com/AI4Bharat/IndicWav2Vec}}, extracting features from layers $7-11$ and projecting them to $80$ dimensions. Data augmentation includes $3$-way speed perturbation and SpecAugment (time warping, frequency masking up to $27$ bins, and time masking for $5\%$ of the sequence). Upstream model parameters remain frozen. Hyperparameters are empirically tuned based on validation performance.

\subsection{Evaluated Methods}
We evaluated the following configurations:

\begin{itemize}
  \item \textbf{Base-ASR}: Standard ASR without DID as an auxiliary task.
  \item \textbf{ASR-DID}: Base-ASR with dialect IDs prepended to text.
  \item \textbf{ASR-DID-SC}: ASR-DID with CTC-based self-conditioning ($2nd$ Conformer layer) \cite{William_LID_ICASSP2023}.
  \item \textbf{ASR-DID-AUX}: ASR-DID with auxiliary CTC for the DID task ($2nd$ Conformer layer) \cite{lonergan2024low}.
  \item \textbf{ASR-DID-ROB}: Similar to the ASR-based DID reported in \cite{ICASSP25_DID}, with the ASR configuration identical to Base-ASR.
  \item \textbf{ASR-BN}: A joint ASR-DID that includes a DID block with a Bottleneck Encoder (see Figure~\ref{fig:bneck_enc}) followed by mean pooling and a linear classifier for dialect prediction (bottleneck dimension: $32$, $4$ attention heads, $0.1$ dropout).
  \item \textbf{ASR-ROB}: Joint ASR-DID using a RoBERTa encoder in the DID block; CTC embeddings derived from the Conformer output are projected through a linear layer before being fed into RoBERTa (hidden: $64$-dim, layers: $2$, heads: $4$).
  \item \textbf{ASR-BN-ROB (Proposed)}: Our proposed method, illustrated in Figure~\ref{fig:proposed_model}, integrates an Attention Encoder with a hidden size of $256$-dim, $4$ attention heads, and a $64$-dim output for DID. The ASR block’s attention module has a hidden size of $1024$-dim, $4$ attention heads, and produces a $256$-dim output. Both blocks utilize two consecutive attention encoder layers to refine representations.
\end{itemize}

\noindent Overview of Trainable Parameters:
\begin{itemize}
  \item Base-ASR, ASR-DID, SC, AUX: $25.32M$ each.
  \item ASR-DID-ROB (baseline): $29.63M$
  \item ASR-BN: $25.45M$
  \item ASR-ROB: $28.91M$
  \item ASR-BN-ROB (Proposed): $31.37M$
\end{itemize}

\noindent \textbf{Dialect ID Requirement:} Only ASR-DID, SC, AUX, and ROB require dialect IDs to be prepended to the training text. The proposed ASR-BN-ROB and its variants (ASR-BN, ASR-ROB) do not.

\section{Results and Discussion}
\begin{figure*}[ht]
    \centering
    \begin{subfigure}{0.48\textwidth}
        \centering
        \includegraphics[width=\linewidth]{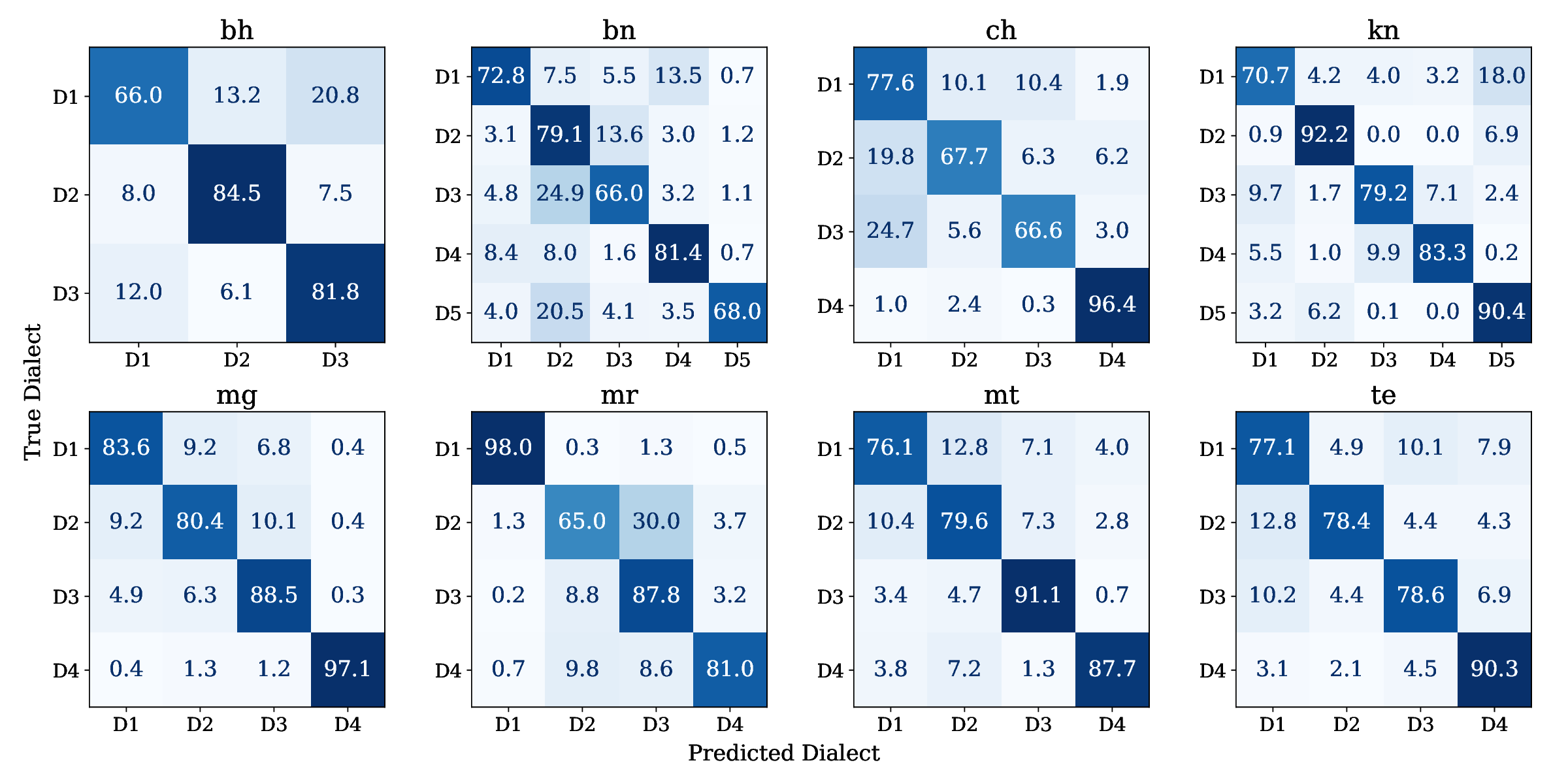}
        \caption{}\vspace{-3mm}
        \label{fig:cm_baseline}
    \end{subfigure}
    \hfill
    \begin{subfigure}{0.48\textwidth}
        \centering
        \includegraphics[width=\linewidth]{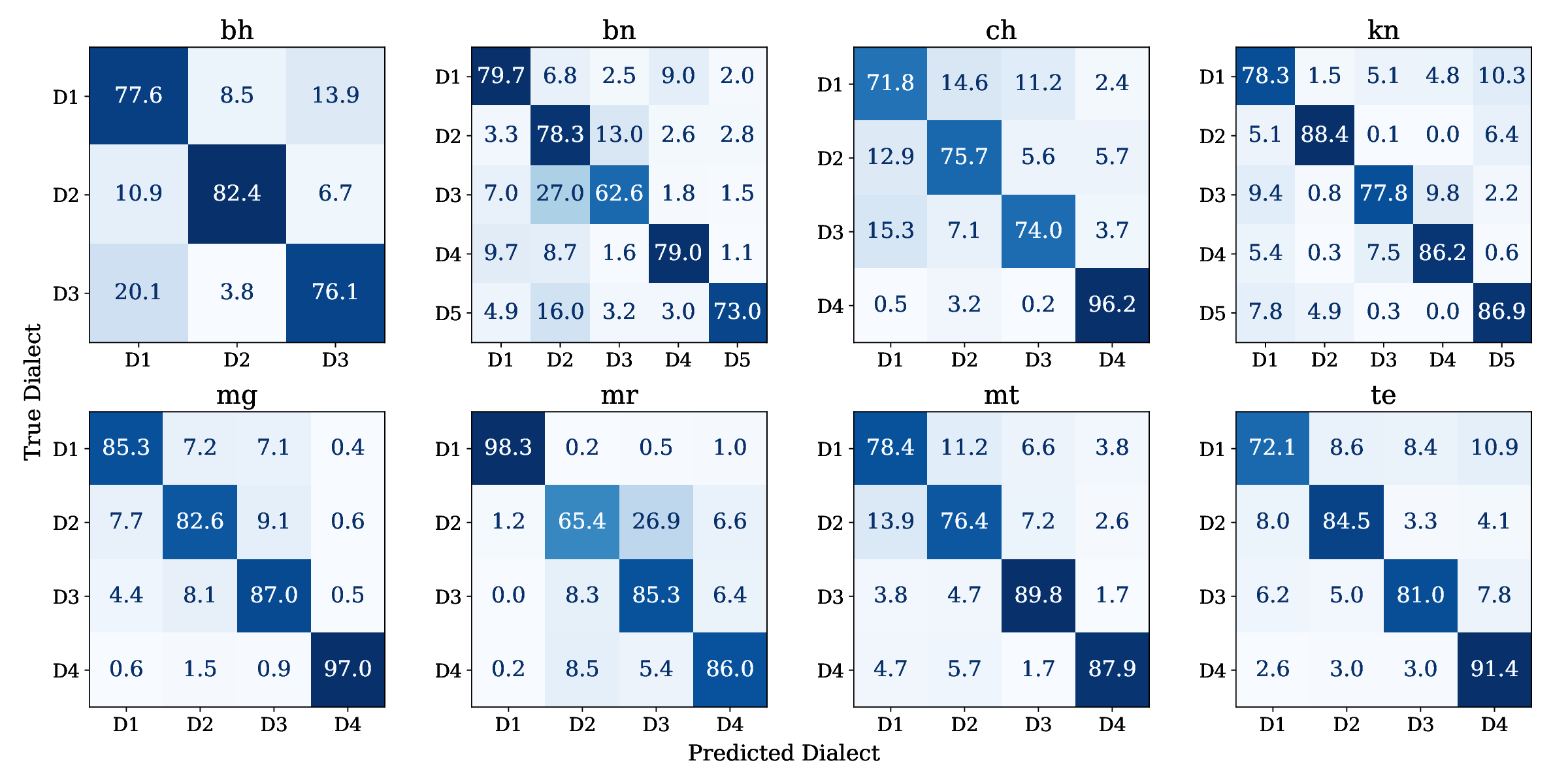}
        \caption{}\vspace{-5mm}
        \label{fig:cm_proposed}
    \end{subfigure}
    \caption{Comparison of confusion matrices for (a) the baseline method (ASR-DID-ROB) and (b) the proposed method (ASR-BN-ROB).}
    \label{fig:cm_comparison}
\end{figure*}

\begin{table}[t]
\centering
\caption{Language-wise DID accuracy.}\vspace{-3mm}
\resizebox{0.90\columnwidth}{!}{%
\begin{tabular}{l|rrrrrrrrr}
\hline
DID Systems & \multicolumn{1}{l}{bh} & \multicolumn{1}{l}{bn} & \multicolumn{1}{l}{ch} & \multicolumn{1}{l}{kn} & \multicolumn{1}{l}{mg} & \multicolumn{1}{l}{mr} & \multicolumn{1}{l}{mt} & \multicolumn{1}{l}{te} & \multicolumn{1}{l}{avg} \\ \hline
ASR-DID & 74.91	& 72.43	& 77.81	& 80.08	& 86.26	& 82.07	& 82.35	& 77.58	& 79.19 \\
ASR-DID-SC & 75.54	& 72.64	& 76.60	& 80.96	& 86.19	& 82.72	& 81.94	& 78.89	& 79.44 \\
ASR-DID-AUX & 75.72 & 73.10 & 76.38 & 81.80 & 86.88 & 81.64 & 82.77 & 80.28 & 79.82 \\
ASR-DID-ROB & 77.43 & 73.38 & 77.00 & 83.18 & 87.31 & 82.90 & 83.67 & 81.06 & 80.74 \\ \hdashline
ASR-ROB & 77.32 & 74.28 & 76.33 & 83.11 & 87.77 & 83.18 & 83.62 & 79.22 & 80.60 \\
ASR-BN & 77.39 & 73.19 & 78.88 & 82.56 & 87.59 & 81.82 & \textbf{84.22} & 80.82 & 80.81 \\
ASR-BN-ROB & \textbf{78.74} & \textbf{74.46} & \textbf{79.38} & \textbf{83.55} & \textbf{87.88} & \textbf{83.66} & 83.11 & \textbf{82.23} & \textbf{81.63} \\ \hline
\end{tabular}%
}
\vspace{-5mm}
\label{tab:did_acc}
\end{table}

\begin{table*}[!h]
\centering
\caption{Language-wise ASR performance.}\vspace{-3mm}
\resizebox{0.90\textwidth}{!}{%
\begin{tabular}{l|rrrrrrrrr|rrrrrrrrr}
\hline
\multirow{2}{*}{ASR Systems} & \multicolumn{9}{c|}{CER} & \multicolumn{9}{c}{WER} \\ \cline{2-19} 
 & \multicolumn{1}{l}{bh} & \multicolumn{1}{l}{bn} & \multicolumn{1}{l}{ch} & \multicolumn{1}{l}{kn} & \multicolumn{1}{l}{mg} & \multicolumn{1}{l}{mr} & \multicolumn{1}{l}{mt} & \multicolumn{1}{l}{te} & \multicolumn{1}{l|}{avg} & \multicolumn{1}{l}{bh} & \multicolumn{1}{l}{bn} & \multicolumn{1}{l}{ch} & \multicolumn{1}{l}{kn} & \multicolumn{1}{l}{mg} & \multicolumn{1}{l}{mr} & \multicolumn{1}{l}{mt} & \multicolumn{1}{l}{te} & \multicolumn{1}{l}{avg} \\ \hline
Base-ASR	& 4.89	& 4.62	& 3.86	& 4.79	& 6.82	& 3.34	& 5.83	& 4.35	& 4.81	& 15.90	& 16.79	& 11.44	& 25.09	& 21.14	& 14.71	& 18.44	& 23.56	& 18.38 \\ \hdashline
ASR-DID & 4.77	& 4.61	& 3.83	& 4.70	& 6.73	& 3.35	& 5.61	& 4.24	& 4.73	& 15.58	& 16.49	& 11.23	& 24.74	& 21.13	& 15.07	& 17.83	& 23.31	& 18.17 \\
ASR-DID-SC	& 4.85	& 4.71	& 3.77	& 4.78	& 6.79	& 3.38	& 5.83	& 4.30	& 4.80	& 15.81	& 16.91	& 11.09	& 24.90	& 20.97	& 15.14	& 18.23	& 23.50	& 18.32 \\
ASR-DID-AUX & 4.91 & 4.58 & 3.80 & 4.76 & 6.71 & 3.32 & 5.59 & 4.15 & 4.73 & 15.84 & 16.62 & 11.14 & 24.49 & 20.82 & 14.75 & \textbf{17.48} & 23.16 & 18.04 \\
ASR-DID-ROB & 4.86 & 4.61 & 3.91 & 4.77 & 6.72 & 3.31 & 5.64 & 4.23 & 4.76 & 15.62 & 16.66 & 11.36 & 24.97 & 20.67 & 14.76 & 17.75 & 23.46 & 18.16 \\ \hdashline
ASR-ROB & 4.81 & 4.55 & 3.94 & 4.73 & 6.68 & 3.28 & 5.70 & 4.25 & 4.74 & 15.40 & 16.39 & 11.35 & 24.68 & 20.62 & 14.80 & 17.88 & 23.02 & 18.02 \\
ASR-BN & 4.76 & \textbf{4.45} & 3.87 & 4.78 & 6.80 & 3.26 & 5.63 & 4.21 & 4.72 & \textbf{15.34} & 16.26 & 11.35 & 24.73 & 20.87 & 14.66 & 17.72 & 23.26 & 18.02 \\
ASR-BN-ROB & \textbf{4.72} & 4.52 & \textbf{3.72} & \textbf{4.68} & \textbf{6.63} & \textbf{3.21} & \textbf{5.58} & \textbf{4.15} & \textbf{4.65} & 15.35 & \textbf{16.20} & \textbf{10.93} & \textbf{24.43} & \textbf{20.37} & \textbf{14.42} & 17.53 & \textbf{22.61} & \textbf{17.73} \\ \hline
\end{tabular}
}
\vspace{-3mm}
\label{tab:asr_cer_wer}
\end{table*}

In this section, we analyze the performance of existing methods and our proposed multimodal feature fusion approach for joint DID and ASR in Indian languages. The evaluation considers DID accuracy, ASR Character Error Rate (CER), and Word Error Rate (WER). We also examine the impact of dialect-informed ASR and the effectiveness of our model. Given the prevalence of compound words in Indian languages, we exclude spaces when computing CERs.

\subsection{Dialect Identification Performance}

Table~\ref{tab:did_acc} presents language-wise DID accuracy for different systems. The dashed line distinguishes ASR-DID methods that prepend dialect IDs from those that learn DID separately alongside ASR. ASR-DID-ROB serves as the baseline, outperforming other existing ASR-DID approaches.

The ASR-ROB and ASR-BN models, which leverage text and audio-based dialectal features respectively, achieve similar performances with accuracies of $80.60\%$ and $80.81\%$ relative to the baseline. On the other hand, our proposed ASR-BN-ROB model, which fuses both modalities, reaches the highest accuracy of $81.63\%$. This model shows significant improvements in Bhojpuri (bh: $78.74\%$ with an increase of $1.43\%$), Bengali (bn: $74.46\%$ with an increase of $1.08\%$), Chhattisgarhi (ch: $79.38\%$ with an increase of $2.38\%$), and Telugu (te: $82.23\%$ with an increase of $1.17\%$). These results underscore the advantages of multimodal fusion for DID.

To further validate our approach, we compare the confusion matrices of the baseline and proposed methods in Figure~\ref{fig:cm_comparison}. Our model not only improves overall DID accuracy but also enhances generalization across dialects. A key indicator is the reduction in the standard deviation of dialect-wise accuracies, averaging a $16.08\%$ decrease across all eight languages. This reduction demonstrates the robustness and consistency of our model in handling dialectal variations more effectively than the baseline.

\subsection{ASR performance}

Table~\ref{tab:asr_cer_wer} reports the language-wise CER and WER across different systems. The first dashed line separates the Base-ASR method from ASR-DID methods, while the second dashed line distinguishes ASR-DID methods that prepend dialect IDs as the first token from our proposed ASR-DID methods, which do not rely on dialect IDs as the first token. All existing ASR-DID methods achieve slightly better ASR performance compared to Base-ASR, indicating that prepending dialect IDs with the text benefits ASR.

Our ASR-BN-ROB method significantly enhances ASR performance in multiple languages by effectively integrating multimodal dialectal features with the ASR encoder output. It achieves an average CER of $4.65\%$ and WER of $17.73\%$ across $8$ languages, surpassing current methods.

\subsection{Impact of Correct vs. Incorrect DID on ASR}

\begin{table}[!t]
\centering
\caption{Overall ASR performance on incorrect and correct DID prediction utterances using the ASR-DID-ROB method.}\vspace{-3mm}
\resizebox{0.90\columnwidth}{!}{%
\begin{tabular}{l|rr|rr}
\hline
\multirow{2}{*}{ASR Systems} & \multicolumn{2}{c|}{CER} & \multicolumn{2}{c}{WER} \\ \cline{2-5} 
 & \multicolumn{1}{l}{Incorrect} & \multicolumn{1}{l|}{Correct} & \multicolumn{1}{l}{Incorrect} & \multicolumn{1}{l}{Correct} \\ \hline
Base-ASR & 5.67	& 4.66	& 20.74	& 17.93 \\
ASR-DID-ROB (baseline) & 6.01 & 4.51 & 21.85 & 17.43 \\
ASR-BN-ROB (proposed) & 5.68	& 4.46	& 20.72	& 17.14 \\\hdashline
Relative difference (\%) & \textbf{5.52}	& \textbf{1.24}	& \textbf{5.17}	& \textbf{1.62} \\\hline
\end{tabular}
}
\vspace{-3mm}
\label{tab:asr_overall}
\end{table}

To further analyze ASR performance differences between Base-ASR, ASR-DID-ROB, and ASR-BN-ROB, we compare utterances with correct and incorrect DID predictions in the baseline. Table~\ref{tab:asr_overall} presents the CER and WER for these cases.

In the baseline method ASR-DID-ROB, dialect IDs are added to the beginning of the text, resulting in significantly worse ASR performance when DID predictions are wrong. For dialects misclassified, the Character Error Rate (CER) is $6.01\%$ and the Word Error Rate (WER) is $21.85\%$, both noticeably higher than the CER of $4.51\%$ and WER of $17.43\%$ for correctly classified dialects. This implies that incorrect DID predictions negatively impact ASR performance. By contrast, the proposed ASR-BN-ROB model addresses this issue by achieving a relative reduction of $5.52\%$ in CER and $5.17\%$ in WER compared to ASR-DID-ROB when dealing with incorrect DID predictions. These findings underscore the importance of accurate dialect recognition for enhancing ASR performance.

\subsection{Discussion}
The results highlight the limitations of ASR-DID-ROB as a baseline. While it improves DID accuracy over other ASR-based DID methods, its ASR performance declines, especially for incorrect DID predictions. In contrast, the proposed ASR-BN-ROB model enhances both DID and ASR performance by leveraging bottleneck and RoBERTa encoders with attention-based fusion for better integration of acoustic and linguistic features. A paired T-test at a $95\%$ confidence interval confirms statistically significant differences in DID accuracy ($t = 2.9609$, $p = 0.0211$), CER ($t = -7.2334$, $p = 0.0002$), and WER ($t = -5.9856$, $p = 0.0006$) across eight languages. These findings demonstrate the effectiveness of multimodal feature fusion for joint DID and ASR in Indian languages.

\section{Conclusion}
We propose a novel multimodal feature fusion approach for joint dialect identification and automatic speech recognition in Indian languages. By integrating a bottleneck encoder on Conformer outputs and a RoBERTa encoder on CTC embeddings, our method improves both DID accuracy and ASR performance. Experimental results demonstrate that our ASR-BN-ROB model surpasses existing ASR-DID approaches, achieving $81.63\%$ DID accuracy, $4.65\%$ CER, and $17.73\%$ WER. Additionally, we analyze the impact of incorrect DID predictions on ASR and show that our model effectively mitigates this degradation. Future work will explore the impact of multimodal fusion on multilingual, multi-dialect ASR in linguistically diverse scenarios, aiming to enhance adaptability across a broader range of dialectal variations.

\section{Acknowledgment}
This work was supported by the RESPIN project, funded by the Bill \& Melinda Gates Foundation. We thank the RESPIN team and our project partner, Navana Tech, for their contributions to data collection.

\bibliographystyle{IEEEtran}
\bibliography{mybib}
\end{document}